\pdfoutput=1
\documentclass[conference]{IEEEtran}
\IEEEoverridecommandlockouts
\usepackage{times}
\usepackage{graphicx}
\usepackage{dblfloatfix}
\usepackage{url}
\usepackage{wrapfig}
\usepackage{booktabs}
\usepackage{multirow}
\usepackage{lineno}
\usepackage{placeins} 
\usepackage{afterpage} 
\usepackage{pgf}
\usepackage[utf8]{inputenc}

\usepackage{bm}
\usepackage{siunitx}

\newif\iffigs
\usepackage{capt-of}  
\usepackage{cuted}
\usepackage[font=small,labelfont=bf]{caption}
\usepackage{subcaption}

\usepackage[fleqn]{amsmath}
\usepackage{float}
\usepackage{listings}
\usepackage{setspace}
\usepackage{mathrsfs}
\usepackage{amssymb}
\usepackage{nicefrac}
\usepackage{algorithm}
\usepackage{algorithmicx}
\usepackage[noend]{algpseudocode}

\usepackage{xcolor}
\usepackage{listings}
\usepackage{textcomp}
\definecolor{backcolour}{rgb}{0.95,0.95,0.92}
\lstdefinestyle{mystyle}{
    backgroundcolor=\color{backcolour},   
    basicstyle=\ttfamily\footnotesize,
    breakatwhitespace=true,         
    breaklines=true,                 
    captionpos=b,                    
    keepspaces=true, 
    keywords={},
    showstringspaces=false,
    showtabs=false,                  
    tabsize=2
}
\usepackage{pythonhighlight}
\usepackage{flushend}

\makeatletter
\newcommand\fs@spaceruled{\def\@fs@cfont{\bfseries}\let\@fs@capt\floatc@ruled
  \def\@fs@pre{\vspace{0.4\baselineskip}\hrule height.8pt depth0pt \kern2pt}%
  \def\@fs@post{\vspace{-0.4\baselineskip}\kern2pt\hrule\relax\vspace{-12pt}}%
  \def\@fs@mid{\kern2pt\hrule\kern2pt}%
  \let\@fs@iftopcapt\iftrue}
\makeatother

\usepackage{lipsum}
\usepackage{changepage}

\usepackage{etoolbox}

\newcounter{algo}
\newcommand{\aln}[2]{\makebox[1.5em][r]{\scriptsize#1:}\hspace{.45em}#2\par}
\newcommand{\alnx}[1]{\makebox[1.5em][r]{}\hspace{.45em}#1\par}
\newcommand{\ind}{\hspace*{1.2em}}
\newcommand{\cmt}[1]{{\footnotesize\itshape\quad$\triangleright$~#1}}
  {\vspace{3.5pt}\hrule height .9pt \end{figure}}

\usepackage[numbers]{natbib}
\usepackage{multicol}
\usepackage[bookmarks=true]{hyperref}

\begin{document}
\title{Leaning Over to Reach Far-Away Objects}

\author{
  \IEEEauthorblockN{
    Allen Devaraj$^{*,1}$,
    William Xie$^{*,1}$,
    and Nikolaus Correll$^{2}$
  }
  \IEEEauthorblockA{
    $^{1}$University of Colorado Boulder \\
    $^{2}$University of Notre Dame \\
  }
  \thanks{$^{*}$Equal contribution.}
}

\maketitle

\begin{abstract}
 We study supported reaching to extend a humanoid robot's workspace by bracing against the environment for stabilization using a straightforward yet capable whole-body model predictive control (MPC) approach, exclusively with local sensing and fiducial markers. While typical solutions for whole body humanoid control aim at avoiding or accommodating contact, contact is explicitly required to reach objects placed on surfaces far out of normal reach. We employ sampling-based MPC for whole-body leaning, bracing, reaching, and recovery on a Unitree H1-2 humanoid. The controller runs in real-time at 34Hz on a laptop, transfers sim-to-real, extends reach by $\approx50$cm beyond the unbraced standing workspace, and sustains the loaded posture for over four minutes, before pose estimation drifts. The controller reaches the commanded target in 89\% of trials and completes the full reach-and-recovery cycle in 81\%, across 54 trials spanning six targets and three table heights. We further characterize why such a small real-time sampling budget is sufficient, finding that temporal persistence of sampled actions and exploration scale strongly govern success, while CEM provides the widest operating region over planning rate and model mismatch. These results establish a real-world baseline for online, contact-rich humanoid manipulation. Code, videos, and full hardware details are available at \url{https://h12leaner.github.io/}

\end{abstract}

\IEEEpeerreviewmaketitle
\section{Introduction} \label{sec:intro}
Recent progress in whole-body humanoid control has increasingly relied on large-scale reinforcement learning or specialized trajectory-optimization and control pipelines, typically with unwieldy motion-capture systems to bypass state estimation noise and drift. These approaches enable remarkable behaviors, but can require task-specific training data, predefined motion references, or specialized optimization machinery \citep{yang2026omniretargetinteractionpreservingdatageneration, ma2025badminton, zhuang2024humanoidparkourlearning, zhang2026learningathletichumanoidtennis, liu2024opt2skill}. 

In parallel, recent work has shown that comparatively simple model-predictive controllers built directly on mature rigid-body simulators can transfer surprisingly well to real legged robots \cite{howell2022predictivesamplingrealtimebehaviour,alvarezpadilla2024realtimewholebodycontrollegged, zhang2025wholebodymodelpredictivecontrollegged}. In this work, we investigate whether the same philosophy extends to multiple-contact planning and control (MCPC) on humanoid platforms for one representative task: supported reaching to far-away objects, which couples manipulation with balance. For example, in industrial battery disassembly \citep{allen2026roboticagenticplatformintelligent, conway2026golemmodularhumanoidautonomy}, many parts are not accessible without leaning into the battery. Humans overcome this challenge quite easily, and can intuitively move beyond their foot-only support region, acquire and load an upper-body support contact, regulate the resulting long-lever posture, accurately and precisely reach a desired target, and subsequently unload and recover \cite{bernstein1967coordination, todorov2002optimal}. These transitions further depend on contact timing and friction with noisy proprioception. All these challenges combined make the task a demanding test of online whole-body control for legged robots.

We present a sampling-based model predictive control framework for whole-body leaning, bracing and reaching, and recovery on a Unitree H1-2 humanoid. The resulting controller transfers sim-to-real, extends the robot's reachable workspace by more than $50$cm beyond unbraced standing, sustains the loaded posture for more than four minutes, and completes $73$\% of maximum-extent reaches and, across six commanded positions and three table heights, $89$\% of commanded reaches. We further study the whole-body stability, target-tracking error, and steady-state jitter of the controller, as well as the component-level contribution of planner costs and four sampling-based planners.

\begin{figure}[t]
\centering
\includegraphics[width=0.99\linewidth]{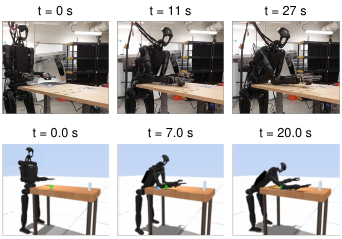}
\caption{Three stages of leaning and reaching Unitree H1-2 in real (top) and simulation (MuJoCo, bottom). With sampling-based model predictive control, the robot can hold the 1.3m reaching stage (frame \textbf{3}), 50cm past the unbraced standing reach of $\approx80$cm for over four minutes with EE accuracy of $2.7\pm0.9$ cm. \label{fig:sketch}}
\end{figure}

\section{Related Work} \label{sec:background}
\paragraph{Whole-body and multi-contact control}
Humanoid whole-body control becomes particularly challenging when employing environmental contacts beyond the feet, a problem commonly studied as multiple-contact planning and control (MCPC), where a controller must reason over both the robot's whole-body motion and a set or sequence of environmental support contacts \citep{ferrari2023multicontact}. Common approaches separate these problems into two: first planning a contact sequence, and then optimizing whole-body control trajectories (and contact forces) under a priori contact constraints.

Several recent systems demonstrate the value of upper-body support contacts for humanoid manipulation. Among ``classical" control methods, most relevant is SEIKO \citep{rouxel2024multicontact}, which performs real-time multi-contact whole-body control on the position-controlled TALOS humanoid, including far-reaching motions that exploit explicitly planned environmental bracing contacts for support. Related work extends this controller with learned multi-support behaviors and automatic support placement for manipulation \citep{rouxel2024flowmatching}, while other approaches adapt contact locations or whole-body posture to improve stability during multi-contact teleoperation \citep{mccrory2025stability}. 
Among learning-based methods which instead encode whole-body behavior into reusable policies or skill spaces, most relevant are: Opt2Skill \citep{liu2024opt2skill}, which learns to track a dataset of dynamically feasible supported reaches generated by trajectory optimization with explicitly-planned contacts (Crocoddyl \cite{mastalli20crocoddyl}), and R$^2$S$^2$ \citep{zhang2025unleashinghumanoidreachingpotential}, which learns a whole-body skill space for large-workspace humanoid reaching, without environmental contacts. 

These prior works have demonstrated large-workspace and multi-contact humanoid reaching with learned and optimized methods. We complement this work with an online whole-body controller that is directly parameterized by geometry and object target commands. We additionally characterize whole-body stability and target-tracking accuracy, commandable workspace, and underlying optimizer operational envelope. 

\paragraph{MuJoCo MPC}
Model predictive control is a common approach for whole-body control of legged robots. MuJoCo MPC (MJPC)~\citep{howell2022predictivesamplingrealtimebehaviour} is a GUI-based platform interfacing physical simulation in MuJoCo, many classes of MPC planners (sampling and derivative-based), and a large set of tasks and robotic embodiments. Recent work has demonstrated MJPC with iterative Linear-Quadratic-Regulator (iLQR) planning on the Unitree H1 humanoid ~\citep{meser2024mujocompchumanoidcontrol,zhang2025wholebodymodelpredictivecontrollegged}, showing the promise of MJPC for real-time, real-world locomotion and manipulation on modern computing platforms. Our work extends MJPC deployment to sampling-based planning on the Unitree H1-2 (two more DoFs) \textbf{without} requiring a motion-capture system.

\section{Methods} \label{sec:method}
\subsection{Model Predictive Control}
We employ MPC within the MuJoCo physics simulator, specifically leveraging MJPC for ease of rapid development and iteration. At each control timestep, the first control $u_0$ is applied and the planner re-solves a trajectory optimization problem from the updated measured state over a finite-horizon $T$ of 1s:
\begin{equation}
\min_{u_{0:T}} \sum_{t=0}^{T} c(x_t, u_t)
\label{eq:mpc_objective}
\end{equation}
subject to dynamics $x_{t+1} = f(x_t, u_t, \Delta t)$, where $x_t \in \mathbb{R}^{n_x}$ is the state (positions and velocities), $u_t \in \mathbb{R}^{n_u}$ is the control (joint torques or position targets), and $c(\cdot)$ is the running cost over the finite horizon. 

\newenvironment{algoblock}[1]{%
  \par\medskip\refstepcounter{algo}%
  \small
  \setlength{\parindent}{0pt}\setlength{\parskip}{1.6pt}%
  \hrule height .9pt \vspace{3pt}
  \textbf{Algorithm \thealgo}\quad #1\par
  \vspace{2.5pt}\hrule height .4pt \vspace{4pt}}%
  {\par\vspace{3.5pt}\hrule height .9pt\medskip}

\begin{figure}
\centering
\begin{minipage}{\columnwidth}

\begin{algoblock}{One planning iteration of sampling-based M(J)PC}
\label{alg:skeleton}
\alnx{\textbf{Input:} control $\theta$, state $x_0$, rollouts $N$, horizon $T$}
\aln{1}{$\bar\theta \leftarrow \textsc{Center}(\theta)$ \cmt{what to sample around}}
\aln{2}{\textbf{for} $k = 1 \dots N$ \textbf{do}}
\aln{3}{\ind $\epsilon^{(k)} \sim \mathcal{N}(0, \Sigma^{(k)})$ \cmt{\textsc{Spread}}}
\aln{4}{\ind $\theta^{(k)} \leftarrow \bar\theta + \epsilon^{(k)}$}
\aln{5}{\ind $u_t \leftarrow \mathrm{spline}(\theta^{(k)}; t)$, roll out $f$ from $x_0$}
\aln{6}{\ind $J^{(k)} \leftarrow \frac{1}{T}\sum_{t<T} c(x_t, u_t)$
    \cmt{Eq.~\eqref{eq:mpc_objective}}}
\aln{7}{\textbf{end for}}
\aln{8}{$\theta \leftarrow
    \textsc{Fold}\big(\{\theta^{(k)}\}, \{J^{(k)}\}\big)$
    \cmt{how the winner is kept}}
\aln{9}{apply $\mathrm{spline}(\theta;0)$; shift $\theta$ forward one step}
\end{algoblock}

\vspace{4pt}

\refstepcounter{table}
\captionof*{table}{\textbf{Table \thetable.}
How planners instantiate Alg.~\ref{alg:skeleton}.
$E$ is the elite set; $k^\star=\arg\min_k J^{(k)}$;
$w_k\propto e^{-J^{(k)}/\lambda}$.
All ablations use $N=20$ rollouts.}
\label{tab:planners}
\centering
\scriptsize
\setlength{\tabcolsep}{2.6pt}
\begin{tabular}{@{}lcccc@{}}
\toprule
Planner & \textsc{Center} & \textsc{Spread} & \textsc{Fold} & $N$ \\
\midrule
Predictive sampling (PS) \cite{howell2022predictivesamplingrealtimebehaviour}
& $\theta$
& $\sigma^2 I$
& $\theta^{(k^\star)}$
& 20 \\

Model predictive path integral (MPPI) \cite{williams2017mppi}
& $\theta$
& $\sigma^2 I$
& $\sum_k w_k\theta^{(k)}$
& 20 \\

Cross-entropy method (CEM) \cite{rubinstein2004crossentropy}
& $\mu_E$
& $\Sigma_E$
& $\mu_E$
& 20 \\

Improved CEM (iCEM)$^\dagger$ \cite{pinneri2020icem}
& $\mu_E$
& $\Sigma_E$, AR(1)
& $\mu_E$
& 20 \\
\bottomrule
\end{tabular}

\vspace{1pt}
{\scriptsize
$^\dagger$iCEM additionally reuses elites between planning iterations.}

\end{minipage}
\end{figure}

MJPC decomposes the cost $c(x, u) = p(l(x,u); R)$, where the base cost is defined as:
\begin{equation}
l(x,u) = \sum_{i=1}^{M} w_i \cdot \text{n}_i(r_i(x, u))
\label{eq:mpc_objective_loss}
\end{equation}

This base cost is a sum of $M$ terms, each comprising a tunable weight $w \in \mathbb{R}_+$, a twice-differentiable norm function $\text{n}(\cdot)$, and residual terms $r \in \mathbb{R}^p$. Then to modulate risk sensitivity, an exponential scalar transformation $p$ is applied for some scalar parameter $R \in \mathbb{R}$, where $R > 0$ is risk-averse and $R < 0$ is risk-seeking.

Each residual $r_i$ captures a specific objective (e.g., reaching a target, maintaining balance, minimizing control effort). The weights $w_i$ are tuned to trade off competing objectives. MJPC implements this through user-defined sensor (residual) specifications in XML with corresponding weight parameters.

MuJoCo models contact and friction with a convex approximation of underlying non-convex, non-smooth contact dynamics. In lieu of analytical derivatives of the soft contact model over a given timestep $x_{t+1} = f(x_t, u_t)$, MuJoCo approximates model derivatives (dynamics Jacobians) via finite differences:
\begin{equation}
\frac{\partial f}{\partial u} \approx \frac{f(x, u + \epsilon) - f(x, u)}{\epsilon}
\end{equation}
A similar expression can be constructed for $\partial f/\partial x$ and finite-difference step $\epsilon$ \citep{zhang2025wholebodymodelpredictivecontrollegged, 6386109}. These state and control Jacobians, and their second-order derivative Hessians, are commonly used for planning in MPC, such as iterative Linear Quadratic Regulator (iLQR) or differential dynamic programming (DDP), but often fail to find local minima (e.g., to discover contact) due to un-differentiable contact discontinuities.

Thus, we primarily explore derivative-free sampling-based planners to solve the task of bracing and reaching. At each replanning step it draws a batch of $N$ candidate control sequences $\{u_{0:T}^{(i)}\}_{i=1}^{N}$, each a perturbation of a nominal spline, rolls every candidate out through the MuJoCo model in parallel, and scores the accumulated cost $\sum_{t} \ell\!\left(x_t^{(i)}, u_t^{(i)}\right)$ along each rollout. Alg. \ref{alg:skeleton} describes one planning iteration of sampling-based MPC, and Tab. \ref{tab:planners} situates four sampling-based planners in Alg. \ref{alg:skeleton}. In short: predictive sampling keeps the single best rollout \cite{howell2022predictivesamplingrealtimebehaviour}, model-predictive path integral (MPPI) forms a cost-weighted average \cite{williams2017mppi}, and cross-entropy method (CEM) and its improved version with colored noise (iCEM \cite{pinneri2020icem}) average a selected elite set and refit the sampling distribution \cite{rubinstein2004crossentropy}. We utilize CEM as the default planner in following experiments and further study the parameters of these planners to understand component-level effects on the task and sim-to-real mismatch in Sec.~\ref{sec:brace_ablation} .

Alternative approaches may use analytical derivatives of rigid-body models (Pinocchio~\citep{carpentier2019pinocchio}) for derivative-based trajectory optimization, such as Crocoddyl~\citep{mastalli20crocoddyl}, which provides feasibility-tolerant DDP solvers. While such methods offer fast convergence through exact gradients, they require more careful contact model specification and may be less robust to model mismatch than derivative-free, sampling-based approaches.  

\subsection{Task Formulation}\label{sec:task}
The task consists of three sequential modes: \textbf{(1)} standing, \textbf{(2)} reaching toward an object, and \textbf{(3)} recovering to standing.

\begin{table}
\centering
\caption{Sampling-controller cost terms, grouped by role. The deployed task defines 39 residuals, grouped here into nine functional families. $[x]_+=\max(0,x)$; $d_i^{\rm seat}$ and $d_i^{\rm keep}$ are clearances to the usable and physical table surfaces; $\mathcal B$ is the selected brace set.}

\label{tab:mjpc_costs_main}
\small
\setlength{\tabcolsep}{3pt}
\begin{tabular}{@{}ll@{}}
\toprule
Group & Representative residual \\
\midrule
Brace          & $\min(\bar d,\,[d_i^{\rm seat}-d_{\rm sat}]_+),\ i\in\mathcal B$ \\
Keep-out       & $[d_{\rm safe}-d_i^{\rm keep}]_+,\ i\notin\mathcal B$ \\
Trunk clear    & $[d_{\rm safe}-d_{\rm trunk}]_+$ \\
Reach          & $p_{\rm hand}-p^{\star}$ \\
Balance        & $\lvert c_{\rm cp}-\Pi_{\mathcal S}(c_{\rm cp})\rvert$ \\
Feet \& stance & foot rise/tilt, capture, knees, hip posture \\
Posture        & $q-q_{\rm home}$; pelvis / torso / waist orientation \\
Height         & $[h_{\min}-(z_{\rm head}-\tfrac{z_L+z_R}{2})]_+$ \\
Regularization & $\dot q;\quad u-q_{\rm home}$ \\
\bottomrule
\end{tabular}
\end{table}

Tab. \ref{tab:mjpc_costs_main} describes the cost terms used for MPC. Here, $c_{\rm cp}=c_{\rm CoM}+0.3\,\dot c_{\rm CoM}$ is the capture point, and $\mathcal S$ is the rectangular support region defined by the feet. Once a selected brace is within 10 mm of the table, it is treated as seated for purposes of extending the support region $\mathcal S$. The quantities $d_i^{\rm seat}$ and $d_i^{\rm keep}$ denote geometric clearances from the usable bracing surface and the physical table surface, respectively; the former uses a $60$mm inset from the table edge to prevent satisfying the seating objective by contacting the lip. The set $\mathcal B$ contains the links selected for bracing by nonzero brace weights. Thus, brace terms encourage those links to approach and seat on the table, while keep-out terms hold all unselected candidate links away from it. We ablate these cost terms in Sec. \ref{sec:brace_ablation}.

The deployed configuration (20 rollouts over a 1s planning horizon) plans at a median 34Hz on a laptop with an AMD Ryzen~AI~9~HX~370 CPU. Preliminary testing on a desktop with an Intel i7-14700F CPU showed $\approx$22Hz planning and increased failure. 

\subsection{State Estimation}
To estimate the floating-base position and velocity $\bm{x}_b,\bm{v}_b\in\mathbb{R}^3$, we fuse proprioceptive measurements, IMU orientation $\bm{R}_b$ and angular velocity $\bm{\omega}_b$, together with joint positions and velocities $\bm{q},\dot{\bm{q}}$ with exteroceptive position measurements from LiDAR and AprilTag fiducials. The proprioceptive estimator uses a random-walk Kalman model for base velocity,
\begin{equation}
\bm{v}_{b,k} = \bm{v}_{b,k-1} + \bm{w}_k,
\qquad
\bm{z}_k = \bm{v}_{b,k} + \bm{\eta}_k,
\label{eq:rwkf}
\end{equation}
where $\bm{w}_k$ and $\bm{\eta}_k$ denote process and measurement noise, respectively.

During standing, each planted foot provides a kinematic observation of base velocity~\cite{bloesch2012} and the table is localized with an AprilTag fiducial marker. Assuming zero world-frame velocity of contact point $k$,
\begin{equation}
\bm{z}_k =
-\Big[
\bm{\omega}_b \times
\big(\bm{p}_k(\bm{q})-\bm{p}_b\big)
+
\bm{R}_b\,\bm{J}_k(\bm{q})\,\dot{\bm{q}}
\Big],
\label{eq:legodom}
\end{equation}
where $\bm{J}_k$ is the translational contact Jacobian. Observations from the two feet are incorporated with standard Kalman updates using contact-weighted, outlier-gated measurement noise~\cite{camurri2017}. Integrating the resulting velocity estimate provides base position, but residual orientation and kinematic errors lead to long-horizon drift \cite{bloesch2012,rotella2014}. We therefore additionally fuse LiDAR- and AprilTag-based position observations to provide an external reference and bound this accumulated position error. 

\begin{figure}[H]
  \centering
  \includegraphics[width=0.85\columnwidth]{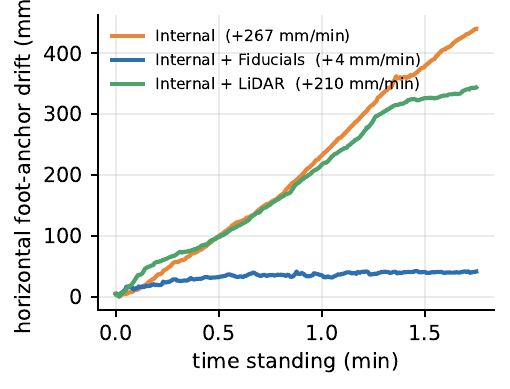}
  \caption{\textbf{Estimator drift during an active stand.}
  While the robot actively holds a stand, Z-axis drift is 0.3mm/min, X-axis drift of the Internal + AprilTag Fiducial estimator is $4$mm oscillation. }
  \label{fig:est_drift}
  \vspace{-8pt}
\end{figure}

\begin{figure}[H]
  \centering
  \includegraphics[width=0.99\columnwidth]{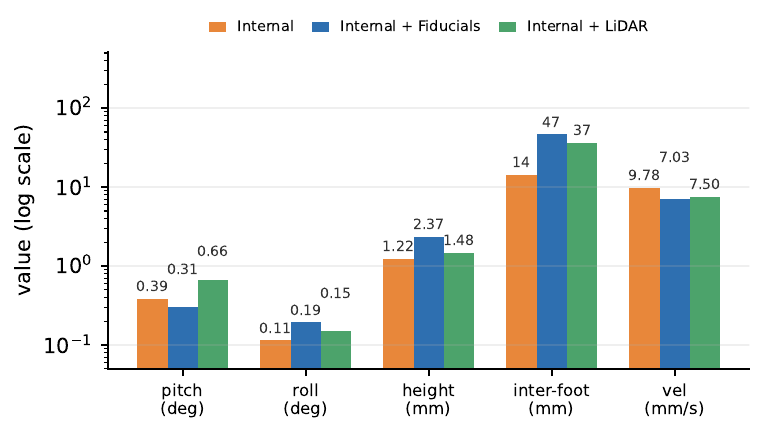}
  \caption{\textbf{Estimated balance states during standing.}
  Base-attitude jitter, height drift, inter-foot wander, and reported velocity
  noise are equivalent across all three estimators.}
  \label{fig:est_balance}
  \vspace{-20pt}
\end{figure}

\section{Experiments} \label{sec:experiments}
We evaluate our methods on a Unitree H1-2 in simulation (MuJoCo) and in the real-world on an adjustable table. Planning and state estimation is done on a laptop, as described in Sec. \ref{sec:task}, with the H1-2 on-board computer publishing camera telemetry. We apply a 1mm thick rubber tape on the table to increase the static coefficient of friction with the H1-2 aluminum body from $\mu\approx0.4$ to $\mu\approx0.8$. 

\begin{figure}[t]
    \centering
    \includegraphics[width=\columnwidth]{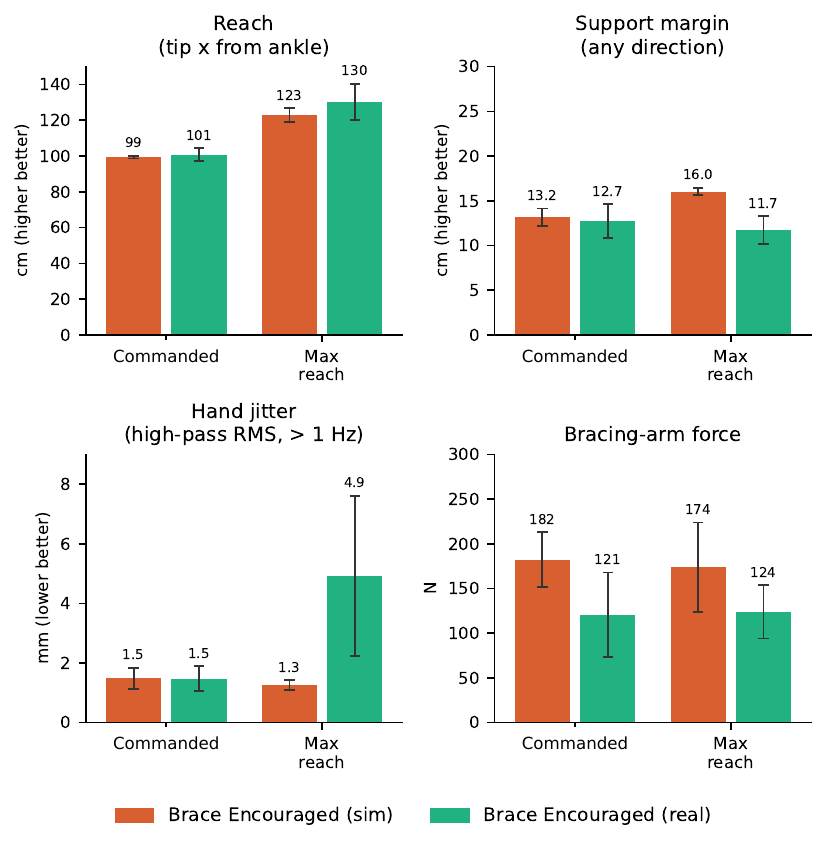}
\caption{\textbf{Reach, stability, hand jitter, and brace loading in successful runs.} Each panel reports two tasks: reaching a commanded target (left pair) and reaching to maximum extent (right pair), in simulation (orange) and on hardware (green). Bars show medians over successful trials; whiskers show the half-ranges. \textbf{Reach} is the gripper-tip distance from the ankle midpoint. \textbf{Support margin} is the minimum margin to the active support-region boundary; \textbf{hand jitter} is high-pass RMS motion above $1$Hz; and \textbf{bracing-arm force} is the measured normal load.}
    \label{fig:brace_ablation}
\end{figure}

\subsection{Estimation}
We evaluate three state estimation methods:
\begin{enumerate}
    \item \textbf{Internal}: IMU + leg-kinematic odometry with kinematic-chain guards,
    \item \textbf{Internal + LiDAR}: the internal estimator fused with FAST-LIO \cite{xu2022fast},
    \item \textbf{Internal + Fiducials}: the internal estimator fused with position observations from an external bundle of four AprilTags (36h11, $5$--$10$cm wide), viewed by the head-mounted Intel RealSense D435i camera.
\end{enumerate}

We evaluate each estimator while the controller actively holds a stand posture. During such a stand the base sways a non-negligible amount, hence, base translation alone cannot distinguish estimator drift from actual motion. We instead use a \emph{foot-anchoring} residual: for a planted foot, forward kinematics from the estimated base pose and measured joint angles should produce a stationary world-frame foot position. Its deviation therefore measures pose error without requiring external ground truth. We report the horizontal and vertical components, along with the inter-foot vector, over the planted-stand interval.

The main difference between the three estimators is horizontal world-frame position (Fig.~\ref{fig:est_drift}). Internal odometry drifts at $+267$ mm/min, or roughly $0.48$ m over a $2$ min hold, while Internal + LiDAR drifts at $+210$ mm/min. The Internal + Fiducial estimator reduces this to $+4$ mm/min, a $\sim60\times$ reduction. The remaining metrics are similar across all three (Fig.~\ref{fig:est_balance}): attitude jitter remains below $0.4^\circ$ in pitch and $0.2^\circ$ in roll, height drift is $1$--$2$ mm, inter-foot variation is $14$--$47$ mm, and velocity noise is $7$--$10$ mm/s. Internal is comparable with the fused variants for attitude and inter-foot consistency, while fiducials primarily improve absolute position. Vertical drift is negligible ($<0.3$ mm over the full trial). We therefore use Internal + Fiducial for the remaining experiments.



\subsection{Leaning, Reaching, and Recovering}
We evaluate two tasks: reaching to a commanded position and reaching to the maximum extent afforded by the planner (max reach). In simulation, the controller completes $90\pm13$\% (18/20) of commanded reach and $80\pm18$\% (16/20) of max-reach trials. On hardware, it completes $89 \pm 8$\% (48/54) of commanded reach trials and $73\pm22$\% (11/15) of max-reach trials. The remainder of this subsection is devoted to stability analysis of successful runs, described in whole by Fig. \ref{fig:brace_ablation}. We investigate failure in greater detail in Sec. \ref{fig:failure}.

\begin{figure}[!t]
\centering
\includegraphics[width=\linewidth]{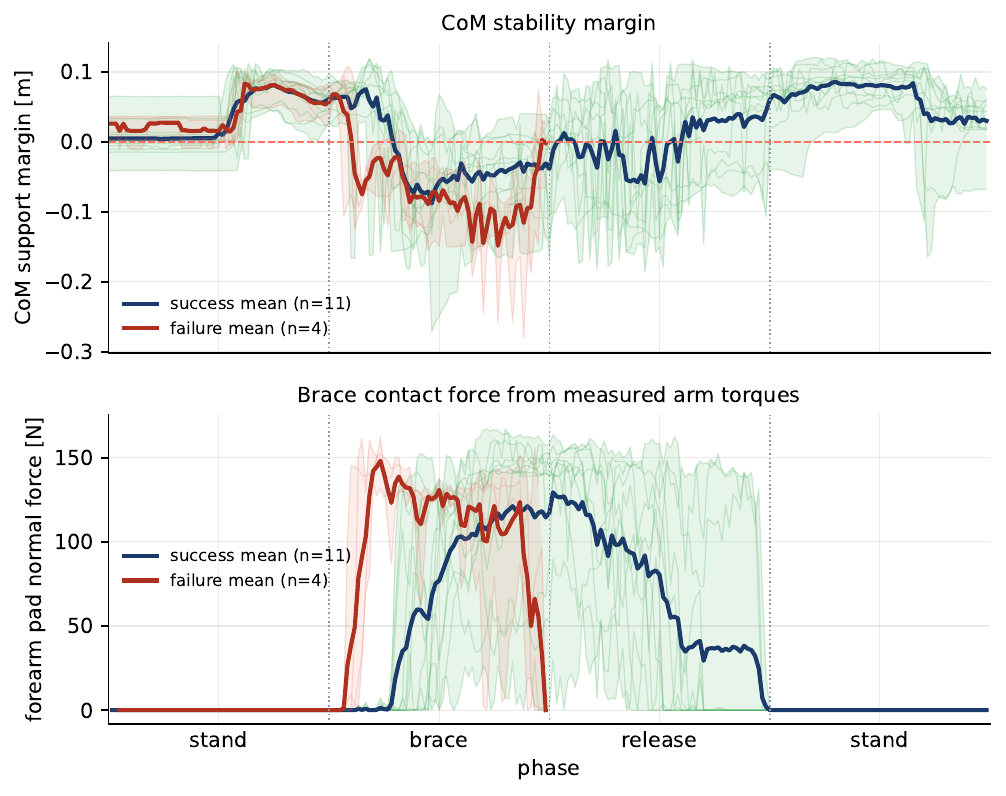}
\caption{\textbf{CoM support margin and reconstructed brace-contact force over hardware trials.}
Time is normalized to the commanded phase boundaries for each trial.
Top: minimum distance from the horizontal CoM projection to the
\textbf{foot-only support polygon}; negative values lie outside the foot
support region. Bottom: normal contact force estimated quasi-statically from
measured bracing-arm joint torques through the contact Jacobian transpose. The
reconstruction gives $4.3$N RMSE against ``ground-truth" force in simulation.}
\label{fig:com_margin_force}
\end{figure}

\begin{figure}[!t]
\centering
\includegraphics[width=\linewidth]{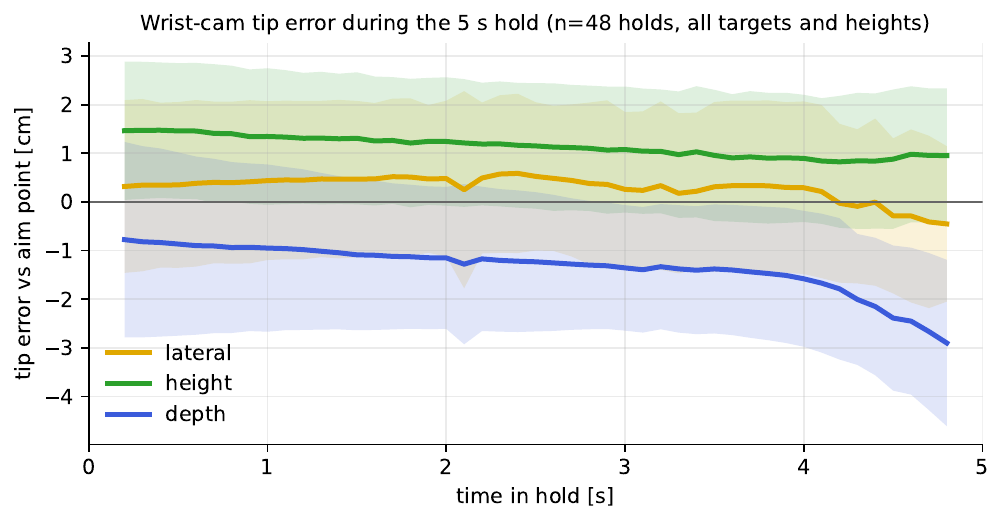}
\caption{\textbf{Gripper tip tracking.} Wrist-camera measured tip error relative to the aim point through the 5s hold, mean $\pm$ 1std over 48 hardware holds across six targets and three table heights. Depth carries a small bias as the hand drives in along the approach ($-1.4 \pm 1.3$\ cm), while the grasp-critical lateral and height axes stay well centered at $+0.2 \pm 1.3$ and $+1.1 \pm 1.3$\ cm within the jaw aperture.}
\label{fig:target}
\end{figure}

\begin{figure*}[t]
\centering
  \includegraphics[width=0.99\linewidth]{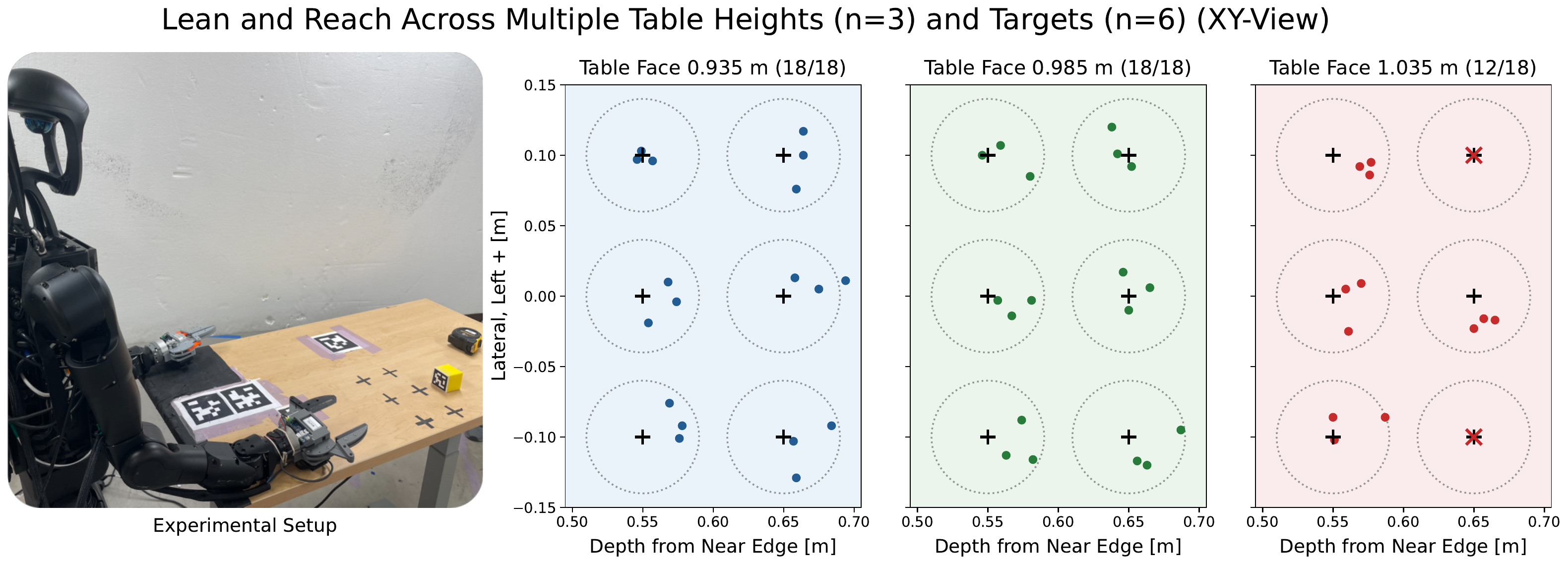}
\caption{\textbf{Commanded reaching across table heights and targets.}
The same controller is evaluated at three table heights and six target positions, with three trials per condition. Dots show wrist-camera measured hold positions in the table plane; crosses denote commanded positions, dotted circles show a 4cm radius, and $\times$ marks conditions where the brace failed before reaching. All targets are completed at the lower and nominal
table heights; at the tallest table, the two deeper lateral targets exceed the current bracing workspace.}
\label{fig:parameterization}
\end{figure*}

\paragraph{Reach} Simulation and hardware reach the same distance. Measured from the ankle midpoint to the gripper tip, a $102$cm commanded reach settles at $99\pm1$cm in simulation and $101\pm4$ cm on hardware. At max reach the tip extends to $123\pm4$cm in simulation and $130\pm10$ cm on hardware, roughly $50$cm beyond the $\approx80$cm the same stance reaches without bracing.

\paragraph{Support margin} Simulation and hardware agree on the stability of the braced posture. The CoM margin is $13.2\pm1.0$cm in simulation and $12.7\pm1.9$cm on hardware at the commanded reach, and $16.0\pm0.4$ versus $11.7\pm1.6$cm at max reach. Successful trials in both conditions keep a positive margin throughout the brace, and the max-reach hardware margin is the smallest because the robot recovers from the longest lever arm.

\paragraph{Bracing-arm force} The simulated controller loads the forearm harder than the real one. Median brace load is $182\pm31$ N at the commanded reach and $174\pm50$N at max reach in simulation, against $121\pm47$ and $124\pm30$N on hardware. Two factors explain the gap: the simulated arm is unclamped, whereas the deployed arm saturates near $100$Nm of shoulder-pitch torque capacity, and the simulated robot leans deeper into the table, which also raises its support margin. Hardware force is reconstructed quasi-statically from bracing-arm joint torques (Fig.~\ref{fig:com_margin_force}).

\paragraph{Hand jitter} High-frequency hand motion during the hold matches across the sim-to-real gap. Jitter above $1$Hz is $1.5\pm0.4$mm RMS in simulation and $1.5\pm0.4$mm on hardware at the commanded reach. At max reach hardware jitter rises to $4.9\pm2.7$mm against $1.3\pm0.2$mm in simulation, consistent with the smaller support margin at that posture. 

\subsubsection{Commanded Position Accuracy}\label{sec:target}
To quantify end effector precision, we command the gripper tip to Cartesian positions in the table frame ($0.65$ and $0.55$m past the near edge, $0.16$m above the surface) and evaluate placement accuracy at convergence. Across $48$ successful commanded reaches in Sec. \ref{sec:icra}, the tip converges to $+2.0 \pm 0.7$cm in the plane transverse to the reach axis, shown in Fig.~\ref{fig:target}, with a lateral component of $+0.2 \pm 1.3$cm and a height component of $+1.1 \pm 1.3$cm. 

\begin{figure}[t]
\centering
\includegraphics[width=\columnwidth]{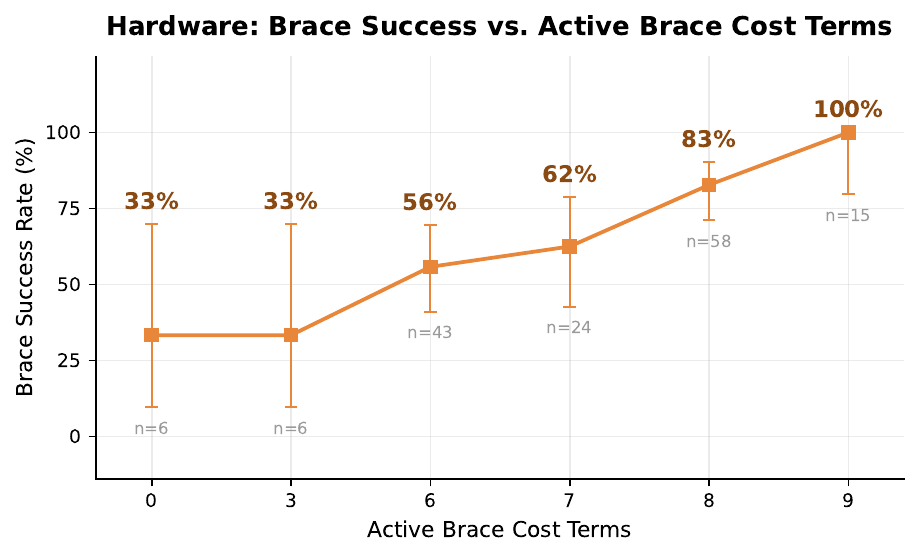}
\caption{\textbf{Brace-cost ablation.} Brace-hold success rate versus the number of active brace cost terms (Wilson $95\%$ intervals; per-point trials $n$
annotated below).}
\label{fig:brace_cost_ablation}
\end{figure}

\subsection{Failure Analysis}\label{fig:failure}

Successful and failed trials separate primarily in how they load and maintain the brace. We analyze the 15 max-reach hardware trials (11 successes, 4 failures). During bracing, both groups move the CoM beyond the foot-only support polygon (Fig.~\ref{fig:com_margin_force}). Successful trials gradually recover to a positive margin up through brace release. In failed trials, the robot slowly falls forward after the initial brace, reaching approximately $-0.14$m overall and below $-0.10$m relative to the front edge before triggering an emergency stop.

The corresponding contact-force profiles further distinguish successful and failing loading strategies (Fig.~\ref{fig:com_margin_force}). Successful trials load the forearm progressively, reaching approximately $125$N on average, with per-trial peaks of $151\pm24$N, and unload gradually during recovery. Failed trials instead develop approximately $150$N near brace onset and subsequently maintain a more variable $100$--$150$N load until collapse. 

Shoulder torque and position measurements also confirm this diagnosis. Successful trials sustain approximately $+10$Nm of shoulder-pitch torque while braced, whereas failures reach $-20$ to $-25$Nm, reflecting configuration ($+0.14\pm0.08$ versus $-0.45\pm0.10$rad). 
To provide some physical intuition, this difference reflects the initial brace placing the shoulder link behind (success) or ahead (failure) of the elbow, and currently is best attributed to sampling variance in the current planner configuration.

\subsection{Table Height and Commanded Position Parameterization}
\label{sec:icra}
We demonstrate that the proposed controller can reach not just one optimized position, but also be commanded by the geometry of the scene. We vary both the table height and commanded reach while otherwise leaving the controller unchanged and report 89\% (48/54) reach and 81\% (44/54) full recovery success across three table heights, $0.935$, $0.985$, and $1.035$m, and six commanded positions arranged across a 20cm by 10cm grid, as shown in Fig.~\ref{fig:parameterization}. The commanded positions are set to, along the forward X-axis, 102cm (10cm inset from the maximum reach) and 92cm, and $\pm10$cm along the Y-axis from the original max reach position. For each table height, the measured table surface is updated in the planner model and the commanded position is then provided directly to the same controller; otherwise, no cost weights or control parameters are changed between conditions. The target is a 5cm cube observed through AprilTags by the wrist-mounted camera (Fig.~\ref{fig:parameterization}, left), which continuously updates the commanded target (visual-servoing). For each individual target, we run three trials consecutively without resetting the robot, with no quantifiable degradation in stability or task performance, for 18 reaches per table height.

The robot completes all $18/18$ holds at $0.935$ and $0.985$m, and $12/18$ at $1.035$m. The two deepest lateral targets at $1.035$m fail during the initial lean, where the forearm brace collapses before the reach is established. Then, at $0.935$m, we report four failures to recover (without an external backwards pull) from the reaching position, similarly due to the forearm brace becoming leveraged too far forward. Across the 48 completed holds, the wrist-camera measurement gives a mean 3D target error of $2.7\pm0.9$cm, with $44/48$ holds within 4cm of the commanded aim point. Mean errors remain similar across table heights: $2.8$, $2.5$, and $3.0$cm, respectively. Figure~\ref{fig:parameterization} shows the corresponding top-down hold locations.

These results suggest that supported reaching as proposed can be used as a manipulation primitive: perception can provide a support height and object pose, and the whole-body controller can realize the corresponding reach without a task-specific motion or controller retuning. We note that planning and modeling improvements, particularly brace pose and contact selection, may further improve performance.

\subsection{Cost and Planner Ablation}\label{sec:brace_ablation}
\paragraph{Brace Cost Ablation}
The intended brace is shaped by several cost terms (the \textit{Brace} group in
Table~\ref{tab:mjpc_costs_main}). Figure~\ref{fig:brace_cost_ablation} ablates these terms \textbf{on hardware}:
brace-hold success (established and sustained brace) rises monotonically with the number of active brace terms, from $56\%$ ($24/43$) with six terms to $100\%$ ($15/15$) with the full nine. While control with zero brace costs and three brace costs exhibit unstable motion with sideways falls, necessitating much fewer trials ($n=6$), we observe emergent bracing behavior and brace success (2/6 for both) that is visually similar to that of explicitly encouraged behavior, as well as rapid recovery.

\begin{figure}[t]
\centering
\begin{subfigure}{\columnwidth}
  \centering
  \includegraphics[width=\columnwidth]{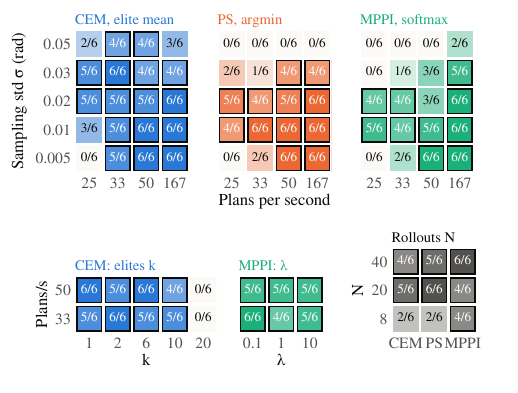}
  \vspace{-20pt}
  \caption{\textbf{Configuration sensitivity.}
  Completion over sampling standard deviation and plan rate for CEM,
  predictive sampling (PS), and MPPI using the hold representation.
  Bottom: CEM elite count $k$, MPPI temperature $\lambda$, and rollout count
  $N$ at 34 plans/s. CEM retains the widest successful noise--rate basin;
  each cell uses six seeds and outlined cells complete at least $4/6$.}
  \label{fig:planner_basin_a}
\end{subfigure}

\vspace{2pt}
\begin{subfigure}{\columnwidth}
  \centering
  \includegraphics[width=\columnwidth]{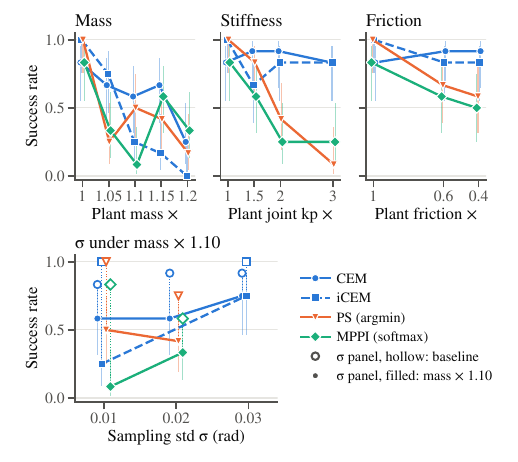}
  \caption{\textbf{Model mismatch.}
  Completion at 34 plans/s when plant mass, joint stiffness, or sliding
  friction differs from the planner model. Bottom: effect of increasing
  sampling standard deviation at nominal mass (hollow) and
  $1.10\times$ mass (filled). CEM is most robust overall, while mass error is
  difficult for every planner. Error bars are Wilson 95\% intervals over 12
  seeds.}
  \label{fig:planner_basin_b}
\end{subfigure}
\caption{\textbf{Sampling-based planner ablation: sensitivity to configuration
and to model error.} Hold spline, $\sigma=0.01$rad, $N=20$, $k=6$ and
$\lambda=1$ unless varied.}
\label{fig:planner_basin}
\end{figure}

\paragraph{Planner Ablation}
One may be curious as to why the proposed controller can be effective while
planning at a relatively slow 34Hz, especially since recent whole-body,
MuJoCo-based MPC systems for legged robots emphasize high-rate control
(100Hz MPPI
\cite{alvarezpadilla2024realtimewholebodycontrollegged}, or 50Hz iLQR with 300Hz smoothing
\cite{zhang2025wholebodymodelpredictivecontrollegged}). To understand the
operating limits of our method, we sweep plan rate, latency, and
planner--plant mismatch \textbf{in simulation}. Broadly, we find that despite the small sampling budget, \(N=20\) rollouts at 34Hz is sufficient because the high-rate joint controller stabilizes the slowly varying posture between MPC updates, and that the planner is most challenged during support transitions. We note that from preliminary testing, derivative-based methods fail much more with the controller as currently constructed and are not pursued further.

For a common $0.01$rad sampling standard deviation and a 1s control trajectory with piecewise-constant, zero-order interpolation of the three control knots (e.g., each sampled action ``knot" or point $u_i...u_K$ in the $K$-knot, $p$-order spline representation of the trajectory is applied, unaltered), all four planners solve the nominal task at 34Hz planning ($4$--$6/6$).
How sampled actions evolve over the 1s rollout strongly affects success: zero-order interpolation holds actions fixed and preserves their effect, whereas linear or cubic interpolation immediately blends toward later samples. Across interpolation order, knot count (trajectory resolution), and noise correlation, configurations whose initial perturbation remains correlated with the executed action for $\gtrsim0.22$s complete $5$--$6/6$ trials, while those decorrelating by $0.20$s complete only $0$--$2/6$. iCEM succeeds with cubic interpolation because its temporally correlated sampling noise preserves the candidate action over this interval. In short, the sampled action must persist long enough for its dynamic effect (contact) to occur in the rollout.

With the trajectory representation fixed, the planners differ through the
\textsc{Fold} update in Alg.~\ref{alg:skeleton}: PS keeps the minimum-cost
rollout, CEM averages the lowest-cost elites and refits their covariance,
and MPPI forms a softmax-weighted average. These update rules produce
different operating ranges as sampling noise and plan rate vary
(Fig.~\ref{fig:planner_basin}a). CEM succeeds in $16/20$ tested
$\sigma$--rate conditions, compared with $12/20$ for PS and $10/20$ for
MPPI. Then, at the nominal $\sigma=0.01$rad, all planners fail at
10Hz planning rate, although standing remains stable even at
5 plans/s. For CEM, increasing the sampling scale partially recovers 10Hz
operation ($4/6$). Latency is less restrictive: at 34Hz, 100ms of uncompensated delay retains $5$--$6/6$ success because new plans still arrive every $30$ms.

Finally, we introduce controlled planner--plant mismatch in mass, joint
stiffness, and sliding friction as a proxy for parametric components of the
sim-to-real gap (Fig.~\ref{fig:planner_basin}b). CEM completes $80/108$
trials, compared with $62/108$ for iCEM, $47/108$ for PS, and $42/108$ for
MPPI. CEM is especially tolerant to stiffness and friction error, while mass
mismatch degrades every planner. Increasing sampling variance also improves
CEM and iCEM tolerance to mass error with little nominal penalty, whereas PS
and MPPI lose nominal successes. Under these settings, CEM exhibits the widest operating range over both plan rate and parametric model error.

\section{Discussion} \label{sec:discussion}
We demonstrate whole-body multiple contact planning and control, e.g. leaning over and reaching far-away objects, with a relatively simple sampling-based MPC framework on a Unitree H1-2 humanoid, sim-to-real. We characterize how our proposed method loads, sustains, and releases an environmental support contact on a H1-2 without external motion-capture technology. 
The controller exposes a surprisingly large commandable workspace using a simple online formulation, but its 81\% full-cycle success also makes clear that online MPC alone does not yet provide the reliability required for autonomous deployment. 

Future work includes further study of brace contact planning, extending the framework to object retrieval for varying payload mass while under a brace, generalizing across support surfaces, inclinations and heights, deploying humanoid skin to improve contact dynamics and contact sensing resolution, exploring GPU-parallelizable and/or learned methods to improve planning rate and quality \cite{liu2024opt2skill}, and predicting, identifying, and recovering safely from failure or anticipated failure. Of note, we observe that a MacBook M5 Pro chip and Intel Ultra 7 265KF CPU can plan at speeds of 60Hz and 45Hz, respectively, on the lean and reach task. Silicon improvements, together with planning and control optimizations, may further facilitate online MPC for such MCPC problems.

\section*{Acknowledgments}
We employ AI tools (Claude Opus 5) to assist with writing code to generate all plots and figures in the manuscript.

\renewcommand{\bibfont}{\footnotesize}
\bibliographystyle{IEEEtran}
\bibliography{references}



\end{document}